# Local-to-Global Self-Supervised Representation Learning for Diabetic Retinopathy Grading

Mostafa Hajighasemlou[a,b*], Samad Sheikhaei[b], Hamid Soltanian-Zadeh[a,c]

[a] *Control and Intelligent Processing Center of Excellence (CIPCE), School of Electrical and Computer Engineering, College of Engineering, University of Tehran, Tehran, 14399, Iran*

[b] *Advancom Lab., School of Electrical and Computer Engineering, College of Engineering, University of Tehran, Tehran, 14395, Iran,*

[c] *Medical Image Analysis Laboratory, Departments of Radiology and Research Administration, Henry Ford Health System, Detroit, MI, 48202, USA*

\* *moshaelen@gmail.com*

## Abstract

Artificial intelligence algorithms have demonstrated their image classification and segmentation ability in the past decade. However, artificial intelligence algorithms perform less for actual clinical data than those used for simulations. This research aims to present a novel hybrid learning model using self-supervised learning and knowledge distillation, which can achieve sufficient generalization and robustness. The self-attention mechanism and tokens employed in ViT, besides the local-to-global learning approach used in the hybrid model, enable the proposed algorithm to extract a high-dimensional and high-quality feature space from images. To demonstrate the proposed neural network's capability in classifying and extracting feature spaces from medical images, we use it on a dataset of Diabetic Retinopathy images, specifically the EyePACS dataset. This dataset is more complex structurally and challenging regarding damaged areas than other medical images. For the first time in this study, self-supervised learning and knowledge distillation are used to classify this dataset. In our algorithm, for the first time among all self-supervised learning and knowledge distillation models, the test dataset is 50% larger than the training dataset. Unlike many studies, we have not removed any images from the dataset. Finally, our algorithm achieved an accuracy of 79.1±1% in the linear classifier and 74.36% in the k-NN algorithm for multiclass classification. Compared to a similar state-of-the-art model, our results achieved higher accuracy and more effective representation spaces.

# 1. Introduction

Diabetic Retinopathy (DR) is a serious retinal disease and is the leading cause of blindness worldwide among individuals aged 25 to 75 [1]. Vision loss, closely linked to diabetes, has become an increasing concern due to the rising prevalence of diabetes mellitus [2]. Over the past few decades, the number of adults aged 20 to 79 with diabetes has surged to an estimated 463 million globally [3]. According to the World Health Organization (WHO), this number is projected to reach 642 million by 2040 [4].

Timely diagnosis and treatment of DR significantly impact both recovery rates and treatment costs. However, despite its importance, many patients are not screened; for example, 40% of diabetic patients in the United States do not undergo screening for DR [5]. The increasing incidence of DR and the challenges associated with timely screening highlight the urgent need for faster and non-invasive diagnostic methods. DR can cause retinal capillaries to bleed and enlarge the venous end of these capillaries due to impaired blood circulation, ultimately resulting in vision loss [6]. Diabetes may develop in the body for years without noticeable symptoms, gradually damaging the retina and vision. This damage to retinal vessels can be detected through simple retinal imaging. Early diagnosis using retinal images and preventing disease progression or initiating treatment in its early stages is a key research focus [7].

Over the past decade, researchers have made significant efforts to apply machine learning and deep learning in medical applications. Recent advances in deep learning, coupled with the development of simulation hardware, have demonstrated the potential of deep learning methods to extract features from diabetic images effectively. Various algorithms have been employed in the field of classification and diagnosis of damaged areas in retinal images. In [8], a supervised learning model consists of two Inception-v3 networks. The first network takes the original image as input and extracts its features. The second network inputs intermediate retinal images, and the neural network extracts features from these two images. Finally, classification is performed based on the feature space extracted from both images.

In [9], it uses Self-Supervised Learning (SSL) to create a pre-trained model and then classifies images into two classes using unsupervised learning. The authors propose a new self-supervised method to improve detecting and segmenting anomalies in medical images. This method includes a novel optimization approach that contrasts one class of normal images with multiple classes of synthesized pseudo-anomalous images, aiming to create a feature space using the features extracted by the proposed algorithm to segment and identify damaged areas.

In [10], researchers have proposed a lesion-aware transformer using weakly supervised learning, implemented through an integrated model composed of a pixel relation-based encoder and a lesion filter-based decoder. The mentioned network, designed for lesion detection localization, utilizes a ResNet-50 network and leverages a weakly supervised learning model that relies on image-level labels. The pixel relation-based encoder is designed to adapt to pixel appearance variations by modeling pixel correlations, while the lesion filter-based decoder is proposed for lesion detection. This involves identifying the presence of lesions, determining their boundaries, and more accurately classifying images based on their features. By using the quadratic weighted kappa, this research achieved an AUC of 98.7% on the Messidor dataset and 89.3% on the EyePACS dataset.

In [11], a self-supervised contrastive learning algorithm used the EyePACS dataset as the training dataset. The neural network is a pre-trained U-Net, which was later fine-tuned on several retinal vessel and lesion segmentation datasets. This article is the first to demonstrate that using contrastive SSL and a pre-trained network can identify blood vessels, the optic disc, the fovea, and various lesions without providing any labels.

In [12] presents a semi-supervised algorithm for classifying retinal images. Smaller labeled datasets can be collected at a significantly lower cost than large, manually labeled datasets. Therefore, the researchers propose a weakly semi-supervised contrastive learning framework for learning representations using smaller labeled datasets. Specifically, a semi-supervised model is trained to generalize labels from a small labeled dataset to a larger unlabeled dataset. The proposed model utilizes ResNet-50 (v2) as the feature-extracting neural network for the OPHDIAT dataset.

Considering the reviewed articles: (1) These studies use traditional convolutional networks as the core of their architecture, and thus, we will use vision transformer (ViT) networks. (2) These studies rely on large datasets to train neural networks, which are not typically available in medical applications. Therefore, we aim to propose an algorithm capable of learning from a small amount of data. (3) The algorithms presented in these articles are not well-compatible with the learning models that have achieved high accuracy in image processing today, making it challenging to apply these algorithms in new models. We will present an innovative model that combines SSL with knowledge distillation (KD), specifically designed to use the EyePACS dataset for training and testing. This approach is unprecedented and has not been explored in any previous research.

## 2. Related Work

Very few works in the SSL and KD fields of study use the EyePACS dataset for training and testing. Most of these works only use this dataset for training. This is because of the challenges this dataset has in the test portion. We review the nearest works that were accomplished for this dataset.

Article [13] introduces a self-supervised fuzzy clustering network. In the first stage, the feature learning module uses multiple convolutional layers to extract features from retinal images. This module ensures that the learned feature representations contain sufficient information from the retinal images for classification. In the second stage, the fuzzy self-supervision module provides supervisory signals to the feature learning module based on the predictions of a fuzzy classification algorithm. This self-supervised strategy leverages the inherent correlations in unlabeled medical images and converts the network's output into fuzzy probabilities corresponding to each class. Finally, for the Messidor dataset, the binary classification accuracy achieved 87.6%.

Article [14] investigates the SimCLR SSL algorithm for DR grading. The proposed method first trains the network using the dataset. Identical images with different augmentations are treated as positive pairs, while images from other classes are considered negative pairs. The network learns to classify positive pairs into the same class and negative pairs into different classes, with the loss function focusing on classification accuracy for these pairs. In this study, the ResNet-50 network was used for both training and testing with a single dataset. The dataset used is a smaller version of the EyePACS dataset. Both the training and testing were done on this dataset, resulting in an accuracy of 77.82%.

In [15] aims to overcome the challenges of acquiring large, labeled datasets necessary for training deep learning models. The study's main contributions can be outlined: First, it introduces a novel approach utilizing KD to transfer the knowledge embedded in a complex, parameter-rich model into a simpler one. Second, it harnesses unlabeled data to impart sufficient knowledge to the simpler model without extensive training. Third, the method's effectiveness is demonstrated by designing a simple, efficient network for analyzing DR images, which can be embedded in medical imaging devices. The process begins by training a robust teacher model on a labeled dataset, generating pseudo-labels for the unlabeled data. The EyePACS dataset was used to train the designed network. This study removed noisy and low-contrast data from the dataset, eliminating approximately 28%. Ultimately, the model's performance was evaluated, achieving a binary classification accuracy of 76.8%.

In [16], the researchers utilized supervised learning with three significant contributions: Firstly, they enhanced and provided better contrast and detail of retinal fundus images related to DR by proposing a new preprocessing method to facilitate disease detection. Secondly, they introduced a novel combination of two ViT and Capsule Network (CapsNet) classification models. Thirdly, they proposed an innovative automatic approach for diabetic nephropathy detection. The modified CapsNet used in this work consists of three sections: The minor hybrid model, primary capsules, and DigitCaps, chosen for their superior ability to retain information about the orientation and location of image features compared to CNN. In this study, the train portion of the EyePACS dataset was used, while the test portion was not utilized. Specifically, 80% of the data was used for training, 15% for testing, and 5% for validation. They ultimately achieved a 78.64% accuracy in multiclass classification.

The review of the aforementioned studies reveals a few key points: First, in the studies examined, the EyePACS dataset is often divided into three sections: training, testing, and, in some cases, validation. However, in some articles, the test data is disregarded. This practice can potentially create issues related to the designed models' reliability, generalization, and robustness. It's important to note that the test section of the EyePACS dataset consists of 53,000 data points compared to 35,000 data points in the training section. Additionally, the test data includes images with more noise, varying contrast, and more significant distortion than the training data. Therefore, classifying the test section is more challenging than the training section.

Second, most of the articles, especially those in the field of SSL, tend to exclude data that the authors consider problematic. It should be noted that this exclusion happens even when using the training dataset in some studies. Moreover, this approach can reduce the model's generalization ability, as it might not be exposed to the full spectrum of data variability during training. Deep learning models have demonstrated the ability to learn important features of images when an appropriate learning approach and model are selected, allowing less significant features to have a reduced impact on the learning process.

Third, the algorithms presented for SSL models do not use advanced and effective methods to control the representation space, so their performance on small or medium-sized datasets is unreliable. With the above points in mind, there is a need for a model that can learn the essential features of the dataset images without excluding any data from the dataset while also ensuring that the model's learning capability is such that it uses less data for training than testing.

The proposed model will achieve this by combining SSL with KD. The model will be designed to learn effectively even when trained on a smaller subset of the dataset, with the expectation that it will still generalize well to a larger and potentially more challenging test set. This contrasts with traditional approaches that rely on large training sets. This combination aims to leverage unsupervised learning and extract essential image features. In addition to accuracy, we will evaluate the proposed model using precision, recall, k-NN, and the self-attention map for the test section of the EyePACS dataset.

## 3. Method

Medical images are highly structured, exhibiting specific, organized patterns, structures, or features that are clinically relevant and important for diagnosis, treatment, or analysis. We employ high-dimensional representation learning, using a combination of SSL and KD models with a transformer-base network, ViT, to analyze medical images for grading and identifying lesion regions.

### 3.1. SSL with KD

Among learning algorithms, we chose SSL for several reasons: (1) Limited labeled data: labeled medical image data is often scarce and expensive to collect. SSL allows us to train models to utilize unlabeled data, which is generally more abundant than labeled data. (2) Unsupervised training: The training phase of self-supervised models does not require labeled data. This unsupervised training leads to the discovery of novel patterns and enhances model interpretability. (3) Interpretable features: SSL can encourage extracting interpretable and clinically relevant features from medical images. This capability can help medical professionals understand and trust the predictions made by AI models. SSL enhances feature learning and generalization, which is crucial for robust performance across various medical imaging modalities.

KD compresses large models into efficient ones, ensuring faster inference and practical deployment in resource-limited settings. KD also improves model robustness and generalization by effectively transferring knowledge from teacher to student models. In this work, we combine SSL and KD. The DINO algorithm [17] employs two networks: a teacher and a student. Unlike traditional KD models, where the student network is derived from the teacher, here the teacher network is dynamically constructed during the learning process using the student network [18]. The student network is trained using unsupervised learning, and the teacher network is built using an Exponential Moving Average (EMA) of the student network [19, 20].

## 3.2. Network

Various works in the fields of KD and SSL can inspire medical applications [21, 22]. Our work leverages techniques and the DINO model to classify DR images, resulting in a model that achieves top-1 accuracy among SSL algorithms for classification. In the proposed method, the teacher is dynamically built during training. Additionally, we can use ImageNet pre-trained weights for both the student and teacher networks. Secondly, the local-to-global learning strategy allows the model to learn detailed features of the images without overfitting. Thirdly, we effectively employ several techniques to prevent the collapse of the representation space produced by both the student and the teacher. Consequently, we achieve superior control over the output feature space compared to other hybrid SSL and KD algorithms. The following sections will elaborate on these advantages.

## 3.3. Backbone

ViT networks are adapted from language processing networks, demonstrating remarkable efficiency in extracting feature spaces for image classification and segmentation [23, 24]. Their unique structure endows these networks with a notable ability to classify medical images effectively.

The ViT demonstrates high efficiency in medical image analysis for several reasons. First, ViT is pre-trained on large-scale image datasets, enabling them to learn fundamental and general representations of images. Second, ViT uses self-attention mechanisms, which allow it to capture long-range dependencies and relationships in medical images. This is especially crucial in tasks like lesion detection, where the spatial context of anomalies is vital. Third, ViT can provide insight into the decision-making process and the damaged area by visualizing attention maps. Fourth, ViT has shown strong generalization abilities across diverse medical image datasets, covering various imaging methods, organs, and diseases. Their ability to adapt to different data distributions makes them highly capable of multiple healthcare applications [25, 26].

The input Images are split into non-overlapping patches. The sizes of patches we use are 16x16 or 8x8. These patches are then passed through a linear layer to form a set of embedding spaces. The ViT network utilizes patch and CLS tokens to enhance feature learning. They incorporated the CLS token to extract complex structures and scatter damaged areas, which is crucial for the classification of retinal images. With its learnability, the CLS token facilitates the creation of a more refined teacher model by effectively capturing and transferring the student network's knowledge [28]. Another significant advantage of the CLS token is its role in the "local-to-global" strategy to help overcome potential limitations of image converters, such as capturing the overall context of images or features that may not be well represented by individual patches alone [29].

We use pre-trained ViT networks on ImageNet, as outlined in Table I [27, 17]. The final projection head of the network consists of a three-layer MLP with 2048 hidden dimensions, L2 normalization, and a fully connected single layer with normalized weights similar to the SwAV [30, 31]. Unlike standard convolutional networks, we do not use any batch normalization.

### 3.4. Preparing Data

Data augmentation significantly enhances the model's ability to learn, generalize, and maintain robustness [32]. However, there are important considerations when using data augmentation. Ensuring label consistency between original and augmented samples is critical to avoid label noise. Additionally, overreliance on aggressive data augmentation methods can lead to overfitting [33]. Therefore, a balanced and specialized approach to data augmentation is necessary. We utilize two global and local image cropping types for teacher and student network learning. Global cropping involves two random crops covering 40% to 100% of the original image. Local cropping, on the other hand, involves six random crops, each covering 5% to 40% of the original image. We train the student network with two global crops and six local crops.

For the teacher network, we only apply the two global crops. Table II provides detailed information about the transformations used to augment the images of global and local crops. This data augmentation strategy is used in BYOL [34] and multi-crop methods [31] with a specific interpolation called bicubic. In ViT networks, bicubic interpolation can be applied to adapt position embeddings when resizing images to different scales. Our work contains KD methods and involves eight crops of original images. Position embeddings in ViT encode the spatial location information of patches in the input image. When the input image size changes, the position embeddings need to be adjusted to match the new scale of the image. Bicubic interpolation helps resize these position embeddings to preserve spatial relationships and smoothness across different scales.

### 3.5. Feature Extraction

As mentioned, our study involves teacher and student networks with the same architecture and structure [35, 18]. We employ two strategies: global and local cropping of images. The image set D1 consists of eight images—two global crops and six local crops—used for training the student network. The D2 image collection includes two global crops and is used for the teacher network. The weights of the student network are referred to as $W_s$, while the weights of the teacher network are referred to as $W_t$. We denote the output of the student network as $O_s$ and the output of the teacher network as $O_t$.

### 3.5.1. Student

Gradient descent is applied solely to the student network, meaning that the learning process targets only the student network. A significant gradient descent slope during the learning process can cause instability, slow convergence, and non-convexity. Gradient clipping is used during training to prevent this effect. If the slope exceeds a predefined threshold, it is reduced to this threshold, which is set to 3 in our research. Stochastic depth helps prevent overfitting and forms robust data representations by encouraging the network to rely on different paths at different times, thereby improving generalization. In our algorithm, 10% of the neural network layers are randomly disabled during the forward learning.

In deep learning, reducing network weights is a regularization technique to minimize overfitting and improve model generalization. Balancing the weight reduction parameter is crucial: too much can hinder the network's learning ability, while too little might have no significant effect. In our study, the weight reduction parameter starts at 0.04 and increases to 0.4, with its increment controlled by the batch size and the total number of epochs desired for the simulation.

Linear warm-up and cosine scheduling techniques have been employed to adjust the learning rate. The primary purpose of modifying the learning rate is to enable the model to escape local minima, thereby enhancing the optimization process. The strategy involves initially increasing the learning rate over the first ten epochs of the simulation, reaching up to 0.0005*batchsize/256. After this warm-up period, the learning rate decreases following a cosine schedule. We use the AdamW optimizer. This optimizer is more adaptive and flexible due to incorporating warm-up rate and cosine schedule strategies for learning rate adjustments. We combined the optimization process with these two techniques, incorporating adaptive learning rate and weight reduction to enhance performance. We refer to the student network as $N_s$.

### 3.5.2. Teacher

In the field of KD, various methods exist for distilling network weights and generating new ones [36]. Dino has demonstrated that the EMA is the most effective method for updating teacher network weights using student network weights [37]. We calculate the trend of the exponential moving average so that the new weights of the network are prioritized over the previous weights, ensuring that fundamental trends significantly impact the network's weight values. This is achieved using a momentum encoder and its update function as described in Eq. (1) [38]. The lambda value is scaled from 0.99 to 1 during the learning process via the cosine function. Consequently, the teacher network weights are updated using the student network weights without applying gradient descent to the teacher network.

During the unsupervised training process for the student model, since student parameters are updated using gradient descent, they may exhibit fluctuations due to gradient noise. EMA provides a more stable and accurate version of the student network's weights for the teacher network, reducing the impact of noisy predictions and resulting in better weights for the teacher network. We refer to the student network as $N_t$.

$$\lambda W_t + (1-\lambda) W_s \rightarrow W_t \quad (1)$$

### 3.6. Network Stabilization

Network collapse refers to a model failing to learn meaningful representations during training, leading to a degenerate solution. Stabilizing the feature space in hybrid models of SSL and KD involves various methods [39]. In Dino, stabilization of the output space is achieved by applying centering and sharpening to $O_t$, and only sharpening to $O_s$. Sharpening adjusts the predicted probabilities to make them more focused or differentiated, enhancing the distinction between predictions of the teacher and student networks. Centering adjusts the teacher's predicted probabilities to remove bias or overconfidence, preventing any one dimension from dominating, though it can increase the risk of collapsing into uniform distributions. Applying both methods to the teacher network balances their effects, preventing collapse and maintaining stability [40]. In this research, the centralization of the teacher network's output is performed with a value C, updated during the learning process according to

Eq. (2). This value C is updated using the EMA for each batch of size B and a constant coefficient m, set to 0.9 as described in Eq. (3). The sharpness is controlled SoftMax function. For the student network, this parameter $\mathcal{T}_s$, is set to 0.1, while for the teacher network $\mathcal{T}_t$, it is set to 0.04, Eq. (4,5).

$$O_t + C \rightarrow O_t \quad (2)$$

$$C \leftarrow mc + (1 - m)\frac{1}{B}\sum_{i=1}^{B} O_t(d_i) \quad (3)$$

$$p_s(x')^{(i)} = \frac{\exp(N_s(x')^i / \mathcal{T}_s)}{\sum_1^{2^{16}} \exp(N_s(x')^i / \mathcal{T}_s)} \quad (4)$$

$$p_t(x)^{(i)} = \frac{\exp(N_t(x)^i / \mathcal{T}_t)}{\sum_1^{2^{16}} \exp(N_t(x)^i / \mathcal{T}_t)} \quad (5)$$

Finally, bypassing the network outputs through the SoftMax function described in Eq. (4,5), we obtain two probability distributions, $p_s$ and $p_t$, each with dimensions $k = 2^{16}$. Our ultimate goal is to minimize the cross-entropy loss between these two distributions with respect to $W_s$ as shown in Eq. (6). It is important to note that, unlike many SSL algorithms such as DINO and SWAV, our algorithm sums the losses for all crops during in each iteration within an epoch. This loss calculation allows the model to better capture the difference between the distributions of the teacher and student networks. Moreover, this technique provides a more reliable slope for SGD and enables more effective centering updates in the teacher's representation space. Additionally, since local crops are applied to both networks, the loss for those crops should not be calculated if the same crops are applied to both the student and teacher networks. We denote the local crops of the image as L and the global crops as G.

$$\sum_{x \in \{G_1, G_2\}} \sum_{\substack{x' \in L+G \\ x' \neq x}} - p_t(x) \log p_s(x') \quad (6)$$

## 4. Experiments

Our research employs three protocols to evaluate the trained network: linear evaluation, k-NN algorithm, and feature space representation. We classify the EyePACS dataset into five classes using a linear classifier and k-NN for classification. To demonstrate the superiority and quality of the feature space extracted by the model, we will visually display this space. This space can provide valuable information about the damaged area and be utilized for image segmentation tasks.

### 4.1. Dataset

We utilized the Kaggle EyePACS dataset, the largest publicly available dataset for classifying DR images, consisting of retinal images collected from American patients. This dataset is divided into a training set comprising 35,126 images and a test set containing 53,576 images. The images were captured using various types and models of cameras. The dataset is annotated into five classes based on the severity of DR: no DR (class 0), mild DR (class 1), moderate DR (class 2), severe DR (class 3), and proliferative DR (class 4) [41]. The EyePACS dataset presents several challenges, including uneven illumination, color distortion, low contrast, motion blur, and the absence of key anatomical features such as the fovea or optic disc [42]. In many studies, particularly those using unsupervised learning techniques, 20% to 25% of lower-quality images are often excluded due to uneven illumination, color distortion, and low contrast [42, 43, 44]. In contrast, our method is designed to retain the entire dataset, allowing the neural network to learn the key features of the images while maintaining robustness against noise and varying image quality. The use of all images, despite their varied conditions, enhances the generalizability of our model.

Unlike traditional SSL networks, where the training dataset typically outweighs the test dataset, our model employs a unique approach where the test dataset is approximately 1.5 times larger than the training set. This uncommon data ratio demonstrates the model's high learning capacity for capturing the critical features of the images. Therefore, our work's most closely related studies have been included compared with other models.

## 4.2. Linear Evaluation

This evaluation protocol assesses the quality of the feature space extracted by the model without fine-tuning the entire feature space on a new dataset. The network's architecture remains consistent between the training and evaluation phases. A linear classification layer is added at the end of the model for evaluation. The output dimensions of this layer correspond to the number of classes in the evaluation dataset. During evaluation, only this linear layer is trained. Images are initially resized to 224 pixels to train this layer, followed by horizontal rotation data augmentation. The training process is supervised, utilizing the SGD optimizer and Cross-Entropy loss function. The learning rate starts at a specified value of 0.001 and is gradually reduced using the CosineAnnealingLR scheduler. After adjusting the learning rate, the model is evaluated on the test data.

Test time data augmentation involves resizing the image to 256 pixels and then cropping it to 224 pixels from the center. The resulting accuracy for the Linear evaluation is shown in Table I. As shown, the top-1 accuracy for the Linear evaluation is 79.1%. We add four CLS tokens to the embedding space sequence for the ViT-S/8 architecture, inspired by feature-based evaluations used in DINO and BERT models [17, 45]. Unlike SSL algorithms like DINO, we do not use an average pooling layer in the feature space before the classification layer. Table III compares the accuracy achieved by our proposed model against other models. This research is the first study that multiclass classification DR on the test portion of the EyePACS dataset using a hybrid SSL and KD network that has been trained only on the training portion of this dataset.

## 4.3. K-NN Algorithm

K-NN is an interpretable algorithm seamlessly integrated into human-in-the-loop systems, allowing medical experts to interact with the model's predictions and provide feedback. This feedback can be used to continuously improve the model's performance and adapt it to specific clinical needs. Our research utilized k-NN cosine similarity classification, particularly effective in high-dimensional feature spaces often encountered in SSL algorithms. Using k-NN with cosine similarity, we can identify data samples close in the feature space, treating these as additional positive training examples. Our k-

NN algorithm does not involve fine-tuning or data augmentation on the test dataset. The best model was the ViT-S/8, so we fixed its weights as parameters and then extracted the feature space of the test images. Finally, we matched the features of an image with its k nearest neighbors from the features extracted by the network during training and the feature space of the extracted test data based on the extracted labels. The resulting accuracy for the k-NN classifier is 74.36%.

### 4.4. Self-Attention Visualizations

The representation of the self-attention space for the CLS token from the last layer of a ViT-S/8 can provide fundamental insights into which parts of the input image are most effective for classification. Displaying this space highlights the algorithm's ability to extract crucial features from other less important ones. Our self-attention maps contain information about the segmentation of an image. Figure I shows the attention space for the CLS tokens of the last layer of the ViT-S/8 network, along with images illustrating all four levels of the disease. However, it is important to note that self-attention blocks, primarily used in natural language processing for sequential data and text, may introduce significant complexities when applied to images. To effectively implement and represent the self-attention space for the CLS token, we need to carefully consider our approach in its implementation, paying particular attention to how the CLS token is handled.

Finally, to evaluate improvements in lesion detection in images, we compared our work with four papers that have demonstrated the best performance in this area. Figure II shows that these studies identified lesion areas covering more significant portions of the image. However, in DR images, lesions are often scattered across different regions, making it crucial to pinpoint the exact location of the lesions. Our model accurately represents lesion locations in the images when comparing Figures I and II. This improvement is attributed to the self-attention mechanism and the hybrid SSL and KD learning techniques we employed.

## 5. Conclusion

In this article, we aim to present an effective model for retinal image classification. We used a hybrid SSL and KD model to achieve this, leveraging the DINO algorithm's proven advantages. By incorporating data augmentation, a ViT backbone, and techniques for stabilizing the representation space, our network achieved the highest accuracy among SSL and KD algorithms for multiclass classification. Finally, we demonstrated the extracted representation space's quality and superiority, highlighting the CLS token's effectiveness. This algorithm, with its combination of self-supervised learning and knowledge distillation, has the potential to be applied to a wide range of medical imaging tasks beyond Diabetic Retinopathy, including tumor detection, organ segmentation, and disease classification.

Table I: ViT networks utilized and their accuracy in multiclass classification for the EyePACS dataset.

| Network | Blocks | Dim | Heads | Tokens | Params | Accuracy (%) |
|---|---|---|---|---|---|---|
| ViT-S/8 | 12 | 384 | 6 | 197 | 21 M | 79.1±1 |
| ViT-S/16 | 12 | 384 | 6 | 785 | 21 M | 77.4±1 |
| ViT-B/8 | 12 | 768 | 12 | 197 | 85 M | 76.6±05 |
| ViT-B/16 | 12 | 768 | 12 | 785 | 85 M | 77±05 |

Table II: Augmentations are applied to both global and local crops of images for the teacher and student network.

| | Data Augmentation |
|---|---|
| **First Global crop** | Random Horizontal Flip, Color Jitter, Random Grayscale, Gaussian Blur. |
| **Second Global crop** | Random Horizontal Flip, Color Jitter, Random Grayscale, Gaussian Blur, Solarization. |
| **Local crops** | Random Horizontal Flip, Color Jitter, Random Grayscale, Gaussian Blur. |

Table III: Comparison of the proposed and S.O.T.A methods in multiclass classification for the EyePACS dataset.

| References | Accuracy (%) | Approach |
|---|---|---|
| [14] | 77.82 | SimCLR SSL-train and test dataset are equal. |
| [16] | 78.64 | Supervised learning- CapsNet, Self-Attention, and CNN. |
| [48] | 83.1 | Supervised learning-80% training and 20% testing CF-DRNet and CNN. |
| [49] | 86.9 | Semi-supervised learning-train and test EyePACS-DeepMT network. |
| **Ours** | 79.1±1 | Hybrid SSL and KD-Test data are 1.5 times bigger than train data Without any data elimination. |

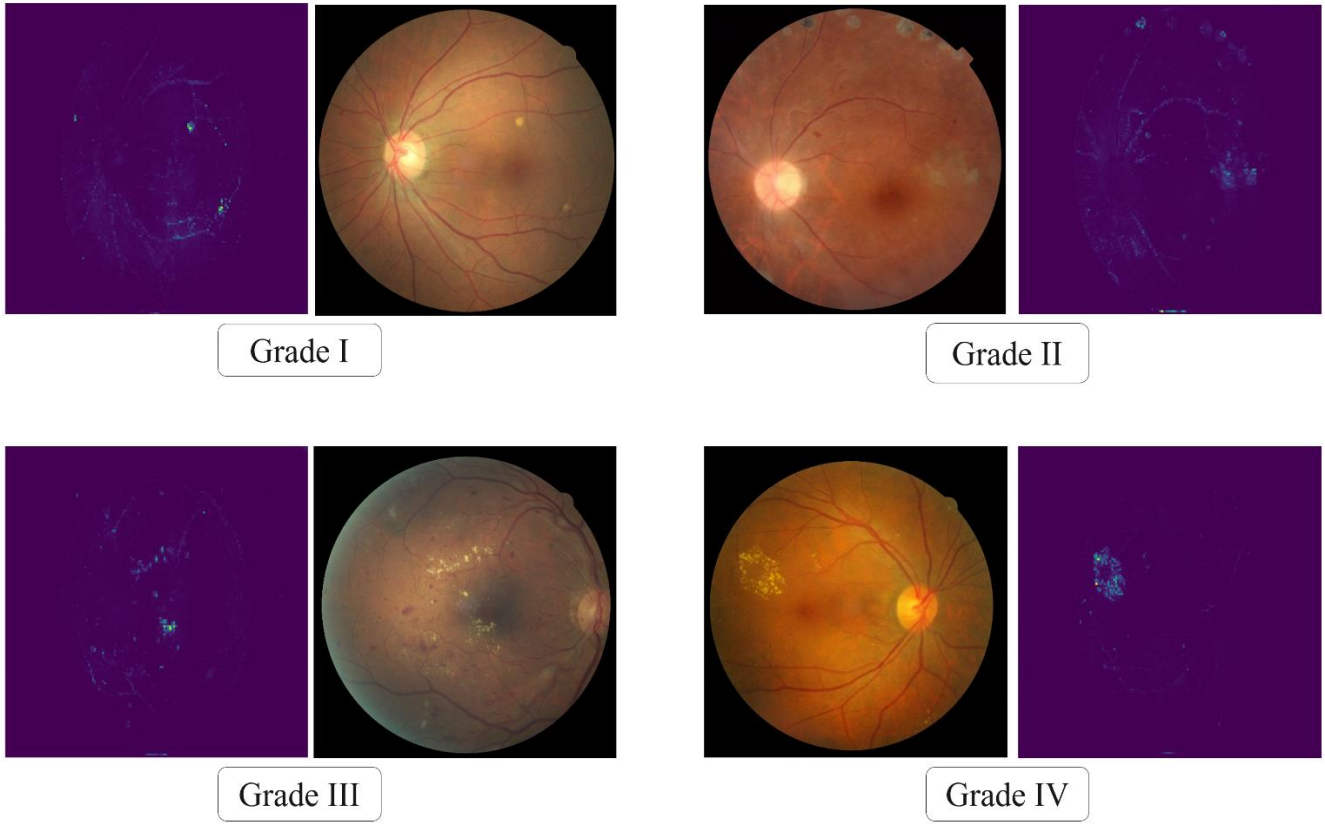

**Figure I:** The self-attention maps of the CLS token in the ViT/8 network with 8x8 patches are used to analyze five grades of DR imaging in the EyePACS dataset, revealing lesion regions in these images.

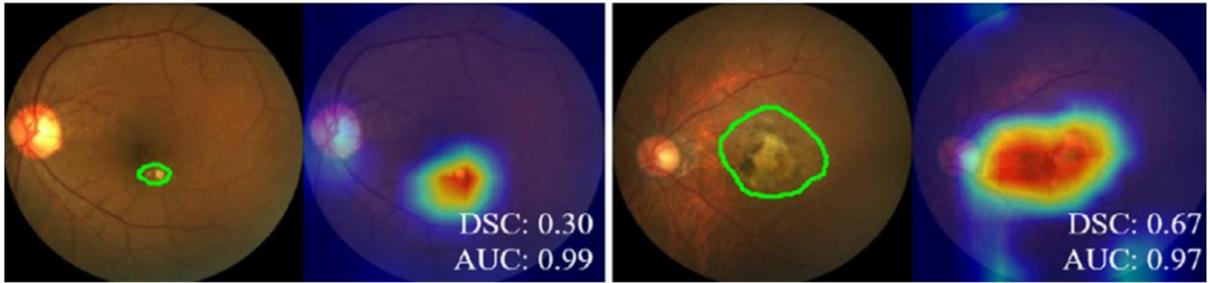

**Gradient-weighted Class Activation Mapping (CAM) [46]**

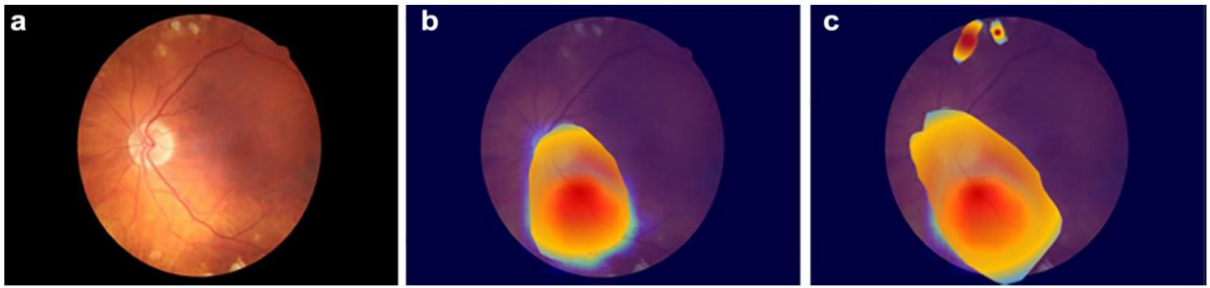

**Self-Supervised Contrastive Learning [42]**

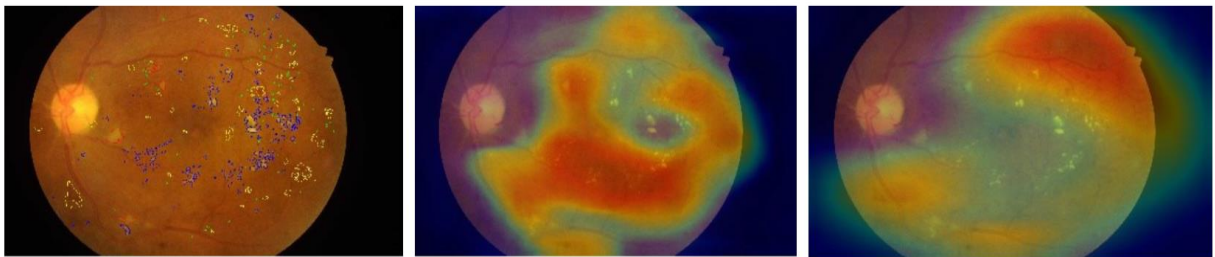

**Ground-Truth [10]**     **LAT[10]**     **CAM[47]**

**Figure II:** Performance of S.O.T.A methods for detecting lesion regions in the EyePACS images dataset.